\documentclass[sigconf]{acmart}

\usepackage{graphicx}

\usepackage{amsfonts,amssymb}
\usepackage[ruled]{algorithm2e}
\usepackage{colortbl} 
\usepackage{xcolor}
\usepackage{subfigure}
\usepackage{amsmath}
\usepackage{mathrsfs}
\usepackage{bm}
\usepackage{epsfig}
\usepackage{rotating, multirow}
\usepackage{epstopdf}
\usepackage{float}
\usepackage{diagbox}
\usepackage{balance}

\newcommand{\tabincell}[2]{\begin{tabular}{@{}#1@{}}#2\end{tabular}}

\AtBeginDocument{%
  \providecommand\BibTeX{{%
    \normalfont B\kern-0.5em{\scshape i\kern-0.25em b}\kern-0.8em\TeX}}}

\begin{document}

\title{Universal Adversarial Perturbations for Vision-Language Pre-trained Models} 

\author{Peng-Fei Zhang, Zi Huang, Guangdong Bai}
\affiliation{%
  \institution{The University of Queensland}
  \country{}}
\email{mima.zpf@gmail.com, huang@itee.uq.edu.au, g.bai@uq.edu.au}

\begin{abstract}

Vision-language pre-trained (VLP) models have been the foundation of numerous vision-language tasks. Given their prevalence, it becomes imperative to assess their adversarial robustness, especially when deploying them in security-crucial real-world applications. Traditionally, adversarial perturbations generated for this assessment target specific VLP models, datasets, and/or downstream tasks. This practice suffers from low transferability and additional computation costs when transitioning to new scenarios.

In this work, we thoroughly investigate whether VLP models are commonly sensitive to imperceptible perturbations of a specific pattern for the image modality. To this end, we propose a novel black-box method to generate Universal Adversarial Perturbations (UAPs), which is so called the \underline{E}ffective and \underline{T}ransferable \underline{U}niversal Adversarial Attack (ETU), aiming to mislead a variety of existing VLP models in a range of downstream tasks. The ETU comprehensively takes into account the characteristics of UAPs and the intrinsic cross-modal interactions to generate effective UAPs. Under this regime, the ETU encourages both global and local utilities of UAPs. This benefits the overall utility while reducing interactions between UAP units, improving the transferability. To further enhance the effectiveness and transferability of UAPs, we also design a novel data augmentation method named ScMix. ScMix consists of self-mix and cross-mix data transformations, which can effectively increase the multi-modal data diversity while preserving the semantics of the original data. Through comprehensive experiments on various downstream tasks, VLP models, and datasets, we demonstrate that the proposed method is able to achieve effective and transferrable universal adversarial attacks.
\end{abstract}

\ccsdesc[500]{Security and privacy}
\ccsdesc[500]{Information systems}

\keywords{Vision-language Pre-training; Universal Adversarial Perturbations; Multi-modal Learning; Transferrable Attack}

\maketitle

\section{Introduction}
Vision-language pre-trained models like CLIP \cite{radford2021learning}, ALBEF \cite{li2021align}, and TCL \cite{yang2022vision}, have emerged as essential tools for understanding intricate relationships between visual and textual elements. These models are pre-trained on large-scale unlabelled datasets and fine-tuned for downstream tasks. Due to their promising performance, they have been widely applied to various vision-language tasks, ranging from information retrieval \cite{liu2021image} to image captioning \cite{zhou2020unified}.

Despite their success, VLP models still face a notable limitation in their ability to withstand adversarial examples, which are crafted by adding imperceptible perturbations to original data. Additionally, when these models are fine-tuned for downstream tasks, the potential vulnerability would also be inherited.

In response to the adversarial vulnerability, several studies have been conducted to assess the adversarial robustness of VLP models \cite{zhang2022towards,lu2023set,wang2023exploring,zhou2023advclip}. For example, Co-Attack \cite{zhang2022towards} learns adversarial examples by enlarging the gap between them with the original paired data in different modalities. SGA \cite{lu2023set} and SA-Attack \cite{he2023sa} adopt data augmentation strategies to increase the input diversity for better disruption of intrinsic cross-modal interactions. However, there are several limitations within previous research. Existing methods typically learn to generate specific adversarial perturbations for each data instance, which might not generalize well to unseen data with different characteristics. In such cases, adversarial perturbations have to be learned from scratch, resulting in extra computational costs. Few studies, such as AdvCLIP \cite{zhou2023advclip}, have started investigating this issue, yet they are either specific to particular models or limited in the imperceptibility. The research problem of \emph{learning universal adversarial perturbations (UAPs) that can be applied to different unseen models, datasets and tasks without additional specialized computation} remains largely open. 

In this paper, we investigate effective and transferrable attacks against VLP models by crafting UAPs for the modality of image. The key challenges of learning effective UAPs are at least three-fold. First, compared to learning sample-specific adversarial perturbations, learning UAPs has to be independent of the specific characteristics of individual samples. Second, conventional universal adversarial attacks in uni-modal cases only need to consider relationships between single instances and their associated labels. In contrast, in multi-modal scenarios, which VLP models focus on, the interactions between different modalities need to be engaged, and their relations are often many-to-many \cite{he2023sa}. Data from different modalities come in different formats and describe the same object from different perspectives, containing diverse and supplementary information. Consequently, complex interactions between different modalities along with the heterogeneity issue cause significant challenges to effective universal adversarial attacks. The third challenge is the transferability. During pre-training, VLP models are customized with various architectures, learning objectives, and even training datasets tailored to specific applications. They could be further fine-tuned in response to different downstream tasks \cite{radford2021learning,li2021align,yang2022vision}. The resulting intrinsic divergence between models makes the learning of transferrable UAPs even more challenging.

To tackle these challenges, we propose a novel black-box UAP generation method, named \underline{E}ffective and \underline{T}ransferable \underline{U}niversal Adversarial Attack (ETU). The ETU focuses on attacking various VLP models without prior knowledge of model details such as architectures, downstream tasks and training datasets. It thoroughly considers the characteristics of UAPs and intrinsic data interactions across different modalities. Specifically, in addition to optimizing the entire space of UAPs, the ETU is designed to improve the utility of the local regions of UAPs, decreasing the interactions between different UAP units. Simultaneous global and local optimizations are expected to enhance the effectiveness of UAPs while boosting their universality. In the meanwhile, a novel data augmentation algorithm named ScMix is proposed, which performs both self-mix and cross-mix data transformations to increase the multi-modal data diversity while preserving the original semantics. Thus, it can help comprehensively exploit cross-modal interactions. UAPs are learned to maximize the dissimilarity between diverse multi-modal data pairs, disrupting cross-modal interactions. As a result, the effectiveness and transferability of UAPs are ensured.

The main contributions of this work are summarized as follows:
\begin{itemize}

\item It is the first attempt to learn UAPs in black-box settings to test the robustness of vision-language pre-trained models. It also characterizes key challenges for launching an effective universal attack in multi-modal scenarios, which can serve as the foundation for future research in this area.

\item A novel effective and transferrable UAP generation method is designed, which improves the utility and transferability of UAPs by comprehensively considering multi-modal interactions. A novel local UAP reinforcement technique and an ScMix data augmentation method are proposed to boost the effectiveness and transferability of adversarial attacks.

\item The proposed ETU is tested on a wide range of VLP models, downstream tasks and datasets, where promising results demonstrate its superiority.   

\end{itemize}

\section{Related work}\label{Rel}

\subsection{Vision-Language Pre-training}
Vision-language models form the cornerstone of a wide range of tasks, e.g., multi-modal retrieval \cite{zhang2021aggregation,zhang2021high}, zero-shot learning \cite{chen2023zero,radford2021learning}, image captioning \cite{chen2015microsoft}, visual question answering \cite{antol2015vqa}, and visual entailment \cite{xie2019visual}. To advance the capability, vision-language pre-training has been introduced. It harnesses large amounts of unlabelled multi-modal data (e.g., image-text pairs) to develop models via self-supervised learning, e.g., multi-modal contrastive learning \cite{li2022blip,radford2021learning,li2021align,yang2022vision}. For instance, the groundbreaking CLIP \cite{radford2021learning} is proposed to learn aligned VLP models that generate embeddings of images and texts with corresponding unimodal encoders. Contrastive learning is utilized to train unimodal encoders by maximizing the similarity between embeddings of matched image-text pairs while simultaneously enlarging the embedding distance between unmatched pairs. BLIP \cite{li2022blip} leverages the dataset bootstrapping method to improve the quality of the training set by synthesizing captions for web images while sieving out noisy captions. Three contrastive learning objectives, i.e., image-text contrastive learning, image-text matching, and image-conditioned language modelling, are utilized to jointly pre-train the model. ALBEF \cite{li2021align} first aligns unimodal representations of image-text pair and then fuses them with cross-modal attention to obtain joint representations. TCL \cite{yang2022vision} utilizes contrastive learning to perform both inter- and intra-modal alignment and preserve mutual information between global and local representations in each modality. Vision-language pre-trained models exhibit outstanding generalizability and transferability in representation learning. Consequently, they have been widely used for downstream tasks through fine-tuning.

\subsection{Adversarial Attack}
\textbf{Conventional adversarial attack}. Adversarial attack aims to mislead target models to make wrong predictions by crafting imperceptible perturbations to inject into original data (i.e., adversarial perturbations)\cite{szegedy2014intriguing,zhang2023proactive,zhang2021privacy}. It is mainly utilized to test the robustness of DNNs. Thereinto, adversarial perturbations can be roughly divided into instance-specific adversarial perturbations \cite{madry2018towards} and universal adversarial perturbations (UAPs) \cite{moosavi2017universal}. As the name implies, instance-specific adversarial perturbations are tailored to particular instances, while UAPs refer to perturbations that are applicable across various instances. Compared to instance-specific adversarial perturbations, UAPs are more applicable for real-world applications as they can deceive different instances of a modal without additional training, even if those instances were not seen during the perturbation crafting process. Representative methods for producing adversarial perturbations include optimization-based methods \cite{szegedy2013intriguing,carlini2017towards}, gradient-based methods \cite{goodfellow2014explaining,madry2018towards} and generative methods \cite{baluja2018learning,poursaeed2018generative,zhang2021proactive}. 

Early attack methods generate adversarial perturbations in the \textit{white-box settings}, where information regarding the victim models, tasks, and data is available. Iterative optimization methods, e.g., I-FGSM, and PGD, that apply multiple-time gradient ascent are utilized for better attack performance  \cite{kurakin2016adversarial,madry2018towards}. However, in real-world applications, it is often the case that target information is often not available, which is referred to as the \textit{grey/black-box setting.} While adversarial perturbations especially UAPs show certain adversarial transferability across different data and models \cite{papernot2016transferability}, such transferability is limited as iterative optimization would incur severe overfitting problems. After some training steps, the perturbations would fit into target models excessively and can hardly transfer to another different model. 

To handle this case, various works have been proposed to enhance the transferability of adversarial perturbations. Taking an ensemble of networks as the target can promote transferability \cite{liu2016delving}. The drawback is that using multiple models comes with computation overhead. Momentum-based methods are designed to accumulate the previous gradients in order to avoid being trapped in poor local maxima \cite{dong2018boosting}. Another line of work proposes to increase the input diversity via data augmentation to prevent overfitting, e.g., data mixing operations \cite{wang2021admix}, data scaling \cite{lin2019nesterov} and random transformations (e.g., resizing, cropping and rotating) \cite{xie2019improving}. 

\noindent\textbf{Adversarial attack against VLP models}. Recently, with the proliferation of VLP models, research has started to investigate the robustness of VLP models. Attacking VLP models is different from those in DNNs. Adversarial attack in DNNs usually concentrates on the classification task. It is a unimodal task that only considers the relation between a single instance and its label, while VLP models focus on multi-modalities. One needs to consider multi-modal data relationships when implementing an attack. For example, Zhang \textit{et al.} \cite{zhang2022towards} propose Collaborative Multimodal Adversarial Attack (Co-Attack) to attack various pre-trained models including CLIP, ALBEF, and TCL. Co-Attack generates multi-modal adversarial perturbations by enlarging the embedding distance between adversarial examples and the original data pairs. As VLP models are usually fine-tuned for downstream tasks, improving the transferability of adversarial perturbations is particularly important to comprehensively evaluate the robustness of VLP models. To this end, Set-level Guidance Attack (SGA) \cite{lu2023set} is proposed to apply data augmentation to increase data diversity and accordingly craft multi-modal adversarial perturbations by minimizing the similarity between them and their matched data from another modality. Self-augment-based transfer attack (SA-Attack) \cite{he2023sa} applies more data augmentations on both original data and adversarial examples to further improve transferability. Co-Attack, SGA and SA-Attack learn multi-modal adversarial perturbations one by one, i.e., first generating adversarial perturbations for one modality and then learning adversarial perturbations for the other. Wang \textit{et al.} \cite{wang2023exploring} propose to learn multi-modal adversarial perturbations simultaneously. 

Despite the progress made, these methods commonly consider data-specific perturbations, suffering from limited generalization ability and transferability. Tailoring perturbations for specific instances makes it challenging to generalize to new data. In addition, it requires additional training for generating perturbations for new data, incurring huge computational costs for large-scale applications. Although AdvCLIP \cite{zhou2023advclip} studies to craft universal adversarial patches, it only concentrates on CLIP. Furthermore, the adversarial patch is not strictly constrained in terms of the perturbation magnitude, making it prone to be identified. To address it, we present the first attempt to learn effective universal adversarial perturbations.

\begin{figure*}
\center
  \includegraphics[width=0.95\textwidth]{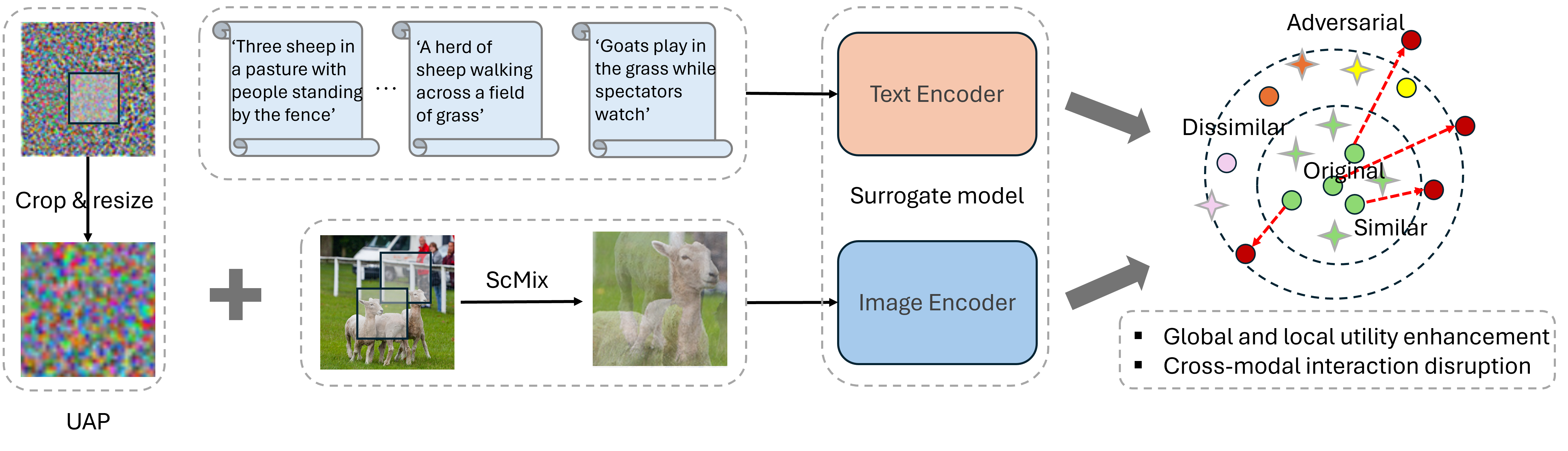}
   \caption{An illustration of the proposed ETU method. The ETU exploits the characteristics of UAPs and diverse cross-modal interactions to improve the utility and transferability of UAPs. Specifically, it generates a variety of similarity-preserving image-text pairs through the ScMix augmentation, which consists of self-mix and cross-mix operations. The ETU optimizes both the entire space and local regions of UAPs by disturbing the similarity between diverse multi-modal data pairs. In light of this, the utility and transferability of UAPs are ensured.}
  \label{fig1}
\end{figure*}

\section{PROPOSED METHOD}\label{METHOD}

\subsection{Preliminaries}

In this work, we aim to learn universal adversarial perturbations that are able to transfer across different VLP models, datasets, and downstream tasks. This means that the VLP models, target datasets, and downstream tasks remain unknown or unavailable during learning. To handle this black-box setting, we train the attack model by utilizing a surrogate dataset and model.

The surrogate multi-modal dataset is denoted as $\mathcal{D}_s = \{({x}_i, {t}_i)\}_{i=1}^n$, where $({x}_i, {t}_i)$ is the image and text pair and $n$ is the number of pairs. For the surrogate VLP model, we denote the image encoder and the text encoder as $f_x$ and $f_t$, respectively. The goal here is to learn a universal adversarial perturbation $\|\delta\|_{\infty} \leq \epsilon$ for the modality of images, where $\epsilon$ is the magnitude of the perturbation and $l_{\infty}$ is the perturbation constraint. The learned UAP can mislead VLP models to wrongly associate images and texts in the target dataset $\mathcal{D}_t$ at the reference time, thus making wrong predictions. For example, in the image-text retrieval, it would make models return incorrect retrieved results. 

\subsection{Overview}
To enable successful black-box attacks, we propose a novel \underline{E}ffective and \underline{T}ransferable \underline{U}niversal Adversarial Attack (ETU) method. As illustrated in Figure \ref{fig1}, the proposed ETU learns UAPs by taking a surrogate model as the victim, which consists of an image encoder and a text encoder. During the learning procedure, the ETU increases the multi-modal input diversity using ScMix data augmentation strategy. UAPs are learned to attack the model by disrupting intra- and inter-modal relationships between original data pairs. Both global and local regions of the UAP would be optimized to improve the effectiveness and transferability.

\subsection{Effective and Transferable Universal Adversarial Attack}
VLP models endeavor to learn the interactions between images and texts for effective multi-modal representation. To achieve successful attacks in multi-modal scenarios, it is necessary to consider multi-modal relationships. An intuitive method is to consider the relationships between the matched pairs, and the UAP is learned by enlarging the embedding distance between the original image and its matched text: 
\begin{equation}\label{eq1-1}
\begin{split}
& \underset{\delta}{\arg\max } \mathcal{L}_{1} = \sum _{i=1}^{n} \left(\ell \left(f_x\left({x}_i + \delta \right),f_x\left({x}_i\right) \right) + \ell \left(f_x\left({x}_i + \delta\right), f_t\left({t_i}\right) \right)  \right),\\
\end{split}
\end{equation}
where $\ell$ is a loss function to quantify the difference between representations of two samples, e.g., the KL-divergence loss.

The objective deters the model from correctly perceiving perturbed images and associating them with their paired texts. Iterative methods such as PGD \cite{kurakin2016adversarial,madry2018towards} are then used to solve the optimization problem given their promising performance. This scheme might work in white-box attack settings, where the target information is available, including target models, data and tasks. However, in real-world applications, such information is largely unknown to the attacker. A common alternative is to utilize a substitute model as the victim target. However, perturbations learned through iterative optimization methods would suffer from low transferability. During the multi-round optimization, perturbations would gradually overfit the victim model. As a result, it can hardly be used to effectively attack another model with different architectures and parameters. This situation will be exacerbated by intrinsic cross-modal interactions in multi-modal application scenarios.

To alleviate these, the ETU leverages two key novel techniques, i.e., local utility reinforcement, and ScMix data augmentation. 

\noindent\textbf{Local utility reinforcement.} This technique enhances the utility of the UAP's local regions for two benefits. First, boosting the local regions can naturally help improve the utility of the entire UAP. Second, it decreases interactions between different local regions, thereby improving the transferability \cite{wang2020unified}. To achieve this, during the UAP learning process, in addition to optimizing the entire UAP, we randomly crop subregions and resize them to the same as the size of the original images. For brevity, we denote this transformation process as $\mathcal{A}_s$. Similar to optimizing the entire UAP, the local regions of the UAP are learned by enlarging the embedding distance between the perturbed images and original pairs:
\begin{equation}\label{eq1-2}
\begin{split}
& \underset{\delta}{\arg\max } \mathcal{L}_{2} = \sum _{i=1}^{n} ( \ell\left( f_x({x_i} + \mathcal{A}_s(\delta) ),f_x({x_i}) \right)\\ 
&  \ \ \ \ \ \ \ \ \ \ \ \ \ \ \ \ \ \ \ \ \ \ \ \ \ +  \ell\left( f_x({x_i} + \mathcal{A}_s(\delta)), f_t(t_i) \right)).\\
\end{split}
\end{equation}

\noindent\textbf{ScMix augmentation.} A representative method for enhancing the transferability is to leverage the data augmentation strategy to increase the input diversity. Learning with diverse inputs can effectively prevent adversarial perturbations from overfitting to specific patterns  \cite{lu2023set,he2023sa}. In addition, in multi-modal scenarios, employing data augmentation can help further exploit the intrinsic cross-modal interactions. In light of these, we design a novel semantic-preserving data augmentation technique. 

\begin{figure}
\center
  \includegraphics[width=0.46\textwidth]{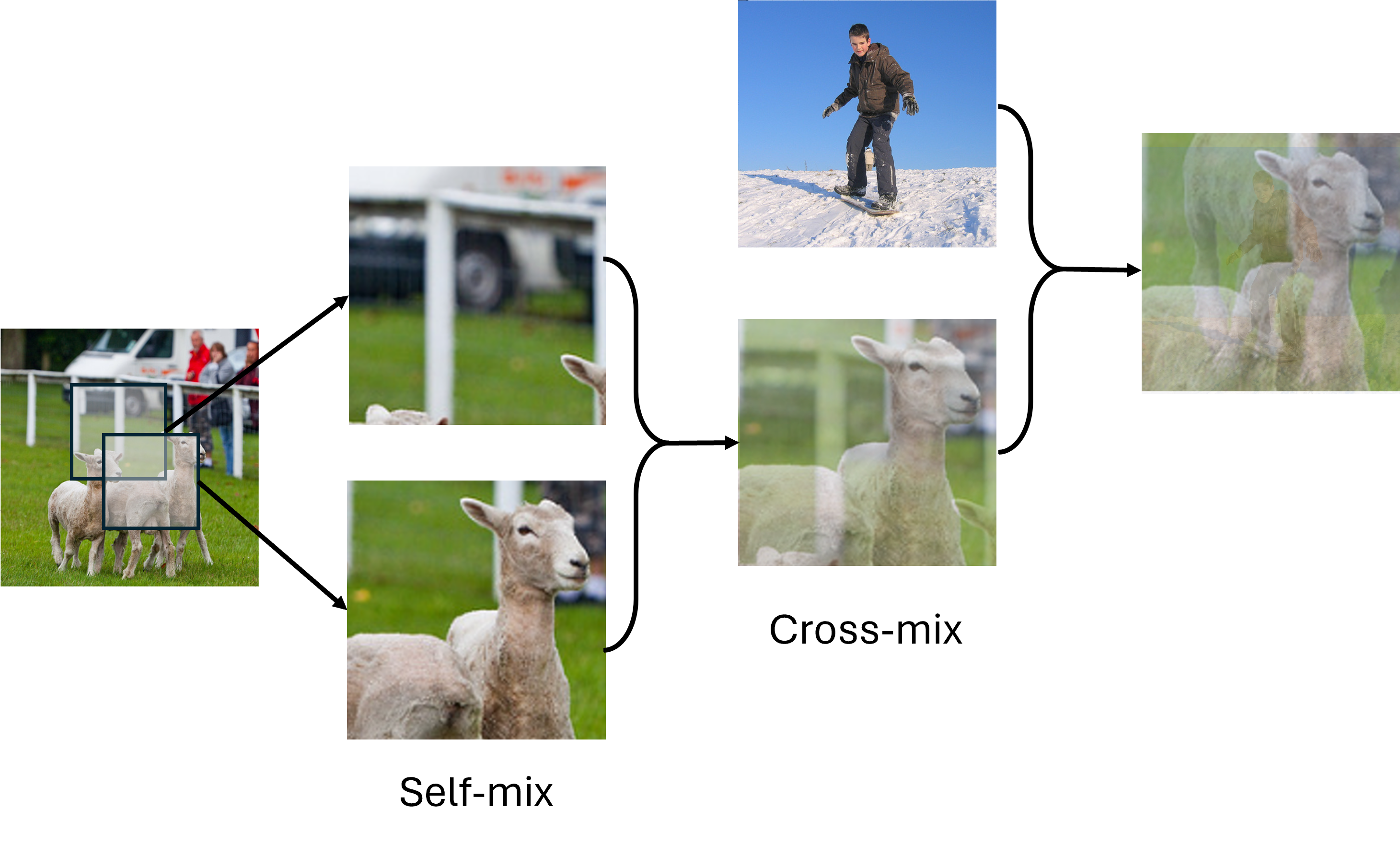}
   \caption{An illustration of the ScMix method, which consists of self-mix operation and cross-mix operations. During self-mix, two local regions of the original image would be randomly cropped and resized to the same size as the original image. Then two rescaled patches would be mixed into a new image. During cross-mix, the self-mixed image would be mixed with another image in a master-slave relation.}
  \label{fig2}
\end{figure}

Specifically, a proper data transformation for augmentation should satisfy the following criteria: diversity and semantic preservation. This means the augmentation should increase the visual difference between the original data and the augmented data. In the meantime, the augmentation should not significantly alter the semantics, which would deter the model from recognizing the data and thus harm the effectiveness of generated UAPs. Under this regime, we propose a novel semantic-preserving ScMix method. As illustrated in Figure \ref{fig2}, the proposed ScMix is a high-order data transformation strategy, which performs self-mix and cross-mix operations to enhance data diversity while preserving semantics. Self-mix constructs new data by mixing up the same original data of different transformations. Specifically, in the self-mix process, two subregions of the original image $x_i$  would be randomly cropped and resized to the same size as the original image. Denote the two rescaled subregions as $x_i^1$ and $x_i^2$. They would be mixed to construct a new data instance $\hat{x}_i$. Finally, the mixed data instance would be interpolated with another data $x_j$ instance that is randomly selected from the mini-batch to craft the final instance $\tilde{x}_i$. Formally, the whole process can be formulated as follows:
\begin{equation}\label{eq2-1}
    \begin{split}
    &p_i = \eta \cdot f_x({x}_i^1) + \left(1-\eta \right) \cdot f_x({x}_i^2),\\
    &\tilde{x}_i =  \underbrace{\beta_1 \cdot \hat{x}_i +  \beta_2 \cdot {x}_j}_\text {Cross-mix},\\
    &\hat{x}_i = \underbrace{\eta \cdot {x}_i^1 + \left(1-\eta \right) \cdot {x}_i^2}_\text {Self-mix},\\
     &s.t. \ \ \eta=\max (\eta^{\prime}, 1-\eta^{\prime}), \ \eta^{\prime}  \sim \operatorname{Beta}(\alpha, \alpha), \\
   \end{split}
\end{equation}
where $\beta_1 > \beta_2 \in [0,1)$ to ensure that the cross-mix would not significantly change the semantics of the original image when injecting the image from another class. $p$ is the prediction for the data after ScMix. $\operatorname{Beta}(\cdot,\cdot)$ represents a Beta distribution. $\alpha \in (0, \infty)$. 

With the ScMix, we can augment the original data to create diverse multi-modal data pairs. This helps effectively prevent overfitting and exploit cross-modal interactions to learn more effective and transferrable UAPs. Moreover, in datasets, e.g., Flickr30K \cite{plummer2015flickr30k} and MSCOCO \cite{lin2014microsoft}, there are often diverse captions to describe an image for different perspectives or language styles. We capture the most matched captions for each image to further increase the multi-modal data diversity. Specifically, denote the caption set for the image $x_i$ is $\{t_{i1}, {t}_{i2}, ..., {t}_{ik}\}$. The augmented multi-modal data pairs are $\{(\tilde{x}_{i1}, t_{i1}, p_{i1}), (\tilde{x}_{i2}, t_{i2}, p_{i2}), ..., (\tilde{x}_{ik}, t_{ik}, p_{ik})\}$, where $\tilde{x}_{i1}, \tilde{x}_{i2}, ..., \tilde{x}_{ik}$ are obtained by applying ScMix on $x_i$. As a result, an enriched dataset would be obtained for learning.

\begin{table}[!t]
\centering
\caption{Attack success rate $(\%)$ on the image-text retrieval task. The CLIP with the ViT-B/16 and Flickr30K are adopted as the source model and dataset for training. The grey background indicates the white-box attack results.}
\label{tab_trans_0}
\centering
\resizebox{8.2cm}{!}{
\begin{tabular}{@{}cc|ccc|ccc@{}}
\toprule
\multicolumn{2}{c|}{Test Dataset}   & \multicolumn{6}{c}{Flickr30K}  \\ \cline{0-7}
\multicolumn{2}{c|}{Task} & \multicolumn{3}{c|}{Image-to-Text}     & \multicolumn{3}{c}{Text-to-Image} \\ \hline
{Target Model} & { Method} & R@1  & R@5 & R@10  & R@1  & R@5  & R@10  \\ \hline
{}  & {ETU$_\text{L}$}  & \cellcolor[HTML]{EFEFEF}{88.96}   & \cellcolor[HTML]{EFEFEF}{76.43}  & \cellcolor[HTML]{EFEFEF}{70.43}  &\cellcolor[HTML]{EFEFEF}{93.49}  &\cellcolor[HTML]{EFEFEF}{88.27} & \cellcolor[HTML]{EFEFEF}{85.32}  \\
{}  &{ETU$_\text{S}$}   & \cellcolor[HTML]{EFEFEF}{{93.13}}   & \cellcolor[HTML]{EFEFEF}{{88.16}}  & \cellcolor[HTML]{EFEFEF}{{83.84}} &\cellcolor[HTML]{EFEFEF}{{96.13}} &\cellcolor[HTML]{EFEFEF}{{93.83}} & \cellcolor[HTML]{EFEFEF}{{92.37}}   \\
\multirow{-3}{*}{\tabincell{c}{CLIP$-\text{ViT-B/16}$}} & {ETU}   & \cellcolor[HTML]{EFEFEF}{88.47}   & \cellcolor[HTML]{EFEFEF}{78.61}  & \cellcolor[HTML]{EFEFEF}{73.07} &\cellcolor[HTML]{EFEFEF}{92.69}  &\cellcolor[HTML]{EFEFEF}{87.5} & \cellcolor[HTML]{EFEFEF}{84.89}   \\\hline\hline
\multicolumn{2}{c|}{Test Dataset}   & \multicolumn{6}{c}{MSCOCO}  \\ \cline{0-7}
{}  & {ETU$_\text{L}$}  & {29.21}   &14.32  & 9.57  & 27.79 & 14.35  & 10.28  \\
{} &{ETU$_\text{S}$}  & {24.65}   & 12.2  & 8.6  & 25.58 & 12.92 & 9.22   \\
\multirow{-3}{*}{\tabincell{c}{ALBEF}} &{ETU}  & {{32.43}}   &{15.76}  & {11.28}  & {29.82 }& {16.25}  & {11.83}    \\\toprule
\end{tabular}}
\end{table}

\noindent\textbf{Cross-modal interaction disruption.} The UAP learned to disrupt the cross-modal interactions by breaking the similarity between diverse paired data in the enriched dataset. Different from current data-specific attack methods that use augmented data to optimize the entire area of UAPs \cite{lu2023set,he2023sa}, we propose to leverage them to enhance the local utility of the UAP. The reason is that UAP is learned over the whole dataset instead of on a single instance as data-specific adversarial perturbation generation. Simply increasing the number of data to optimize the entire UAP would encourage it to overfit the victim model. Instead, utilizing augmented data to optimize local regions of the UAP can further encourage the effectiveness of the UAP while relieving the overfitting problem. Table \ref{tab_trans_0} lists some black-box results, where we can observe that ETU$_\text{S}$ that leverages the ScMix to optimize the entire area of the UAP achieve worse performance than that does not use the ScMix, i.e., ETU$_\text{L}$. And the ETU that uses ScMix to optimize the local utility of the UAP can achieve better transferrable attacks.    

In light of this, the objective of learning UAP based on ScMix is:
\begin{equation}\label{eq2-2}
\begin{split}
&  \underset{\delta}{\arg\max } \mathcal{L}_{3} = \sum _{i=1}^{n_t} \left(\ell\left( f_x(\tilde{x}_{i} + \mathcal{A}_s(\delta)), p_{i} \right) \right.\\ 
& \left. \ \ \ \ \ \ \ \ \ \ \ \ \ \ \ \ \ \ \ \ \ \ \ \ \ \ + \ell\left( f_x(\tilde{x}_{i} + \mathcal{A}_s(\delta) ), f_x({x_i}) \right) \right.\\
& \left. \ \ \ \ \ \ \ \ \ \ \ \ \ \ \ \ \ \ \ \ \ \ \ \ \ \ + \ell\left( f_x(\tilde{x}_{i} + \mathcal{A}_s(\delta)), f_t({t_i}) \right) \right),\\
\end{split}
\end{equation}
where $n_t$ is the caption number.

The overall objective for learning UAP is defined as:
\begin{equation}\label{eq3-1}
 \begin{split}
   &  \underset{\delta}{\arg\max } \mathcal{L} = \mathcal{L}_{1} + \mathcal{L}_{2} + \mathcal{L}_{3},\\
 \end{split}
\end{equation}
where data in $\mathcal{L}_{1}$ and $ \mathcal{L}_{2}$ would also be augmented with different captions for each image.

To solve the optimization problem, we leverage the commonly used projected gradient descent (PGD) \cite{madry2018towards}. The detailed algorithm is summarized in \textbf{Algorithm \ref{alg1}}.

\begin{algorithm}[t!]
       \caption{Universal Adversarial Perturbations for Vision-Language Pre-trained Models}
       \label{alg1}
       \textbf{Require}: Training data $\mathcal{D} = \{({x}_i, {t}_i)\}_{i=1}^n$, mini-batch size $m$, iteration times $T$,  parameters $\epsilon, \alpha, \beta_1, \beta_2$; \\
        \textbf{Require}: Randomly initialize $\delta$;\\
        // Training;\\
        // Exploit diverse matching captions for each image to augment the dataset;\\
        \For{$epoch = 1 \rightarrow T$}{
            $\{({x}_i, {t}_i)\}_{i=1}^l \sim \mathcal{D}$ //  Sample mini-batch from the dataset;\\
            $\{(\tilde{x}_i, t_i)\}_{i=1}^l \leftarrow \{(x_i, t_i)\}_{i=1}^l$ //  Augment each image-text pair via ScMix;\\
           $\mathcal{A}_s(\delta)$ // Randomly crop and resize the UAP;\\
           $ \delta \leftarrow$ Eq.(\ref{eq3-1}) // Update the UAP by optimizing Eq.(\ref{eq3-1});\\
        }
        return $\delta$\;
\end{algorithm}

\section{Experiments}\label{Exp}

\subsection{Settings}
\begin{table*}[!t]
\center
\caption{Attack success rate $(\%)$ on the image-text retrieval task. The CLIP with the ViT-B/16 is adopted as the source model for training. The grey background indicates the white-box attack results. \textbf{Bold} indicates the best results.}
\label{tab_trans_1}
\centering
\begin{tabular}{@{}c|c|ccc|ccc|ccc|ccc@{}}
\toprule
\multicolumn{2}{c|}{Test Dataset}  & \multicolumn{6}{c|}{Flickr30K}     & \multicolumn{6}{c}{MSCOCO}  \\ \cline{0-13}
\multicolumn{2}{c|}{Task} & \multicolumn{3}{c|}{Image-to-Text}            & \multicolumn{3}{c|}{Text-to-Image}  & \multicolumn{3}{c|}{Image-to-Text}     & \multicolumn{3}{c}{Text-to-Image} \\ \hline
{Target Model} & {Method} & R@1  & {R@5}  & {R@10}  & {R@1}  & R@5 & {R@10} & R@1  & R@5 & R@10  & R@1  & R@5  & R@10  \\ \hline

{} & {UniA}  & \cellcolor[HTML]{EFEFEF}{91.9}     & \cellcolor[HTML]{EFEFEF}{82.87}    & \cellcolor[HTML]{EFEFEF}{78.66}  & \cellcolor[HTML]{EFEFEF}{91.14} & \cellcolor[HTML]{EFEFEF}{81.65} & \cellcolor[HTML]{EFEFEF}{79.96}     & {95.5}   & 91.42  & 88.82  & 94.24  & 90.11 & 87.65   \\
{} & {MulA} & \cellcolor[HTML]{EFEFEF}{92.02}   & \cellcolor[HTML]{EFEFEF}{82.04}  & \cellcolor[HTML]{EFEFEF}{76.62}    & \cellcolor[HTML]{EFEFEF}{94.85}  &\cellcolor[HTML]{EFEFEF}{90.42} & \cellcolor[HTML]{EFEFEF}{86.22}   & {95.5}   & 91.6  & 88.87    &96.13  &93.5   &91.76   \\
{}  & {ETU$_\text{L}$}  & \cellcolor[HTML]{EFEFEF}{88.96}   & \cellcolor[HTML]{EFEFEF}{76.43}  & \cellcolor[HTML]{EFEFEF}{70.43}  &\cellcolor[HTML]{EFEFEF}{93.49}  &\cellcolor[HTML]{EFEFEF}{88.27} & \cellcolor[HTML]{EFEFEF}{85.32}    & {93.78}   & 88.85  & 85.17   &95.34  &92.21   &90.49   \\
{}  &{ETU$_\text{S}$}  & \cellcolor[HTML]{EFEFEF}{\textbf{93.13}}   & \cellcolor[HTML]{EFEFEF}{\textbf{88.16}}  & \cellcolor[HTML]{EFEFEF}{\textbf{83.84}}   &\cellcolor[HTML]{EFEFEF}{\textbf{96.13}} &\cellcolor[HTML]{EFEFEF}{\textbf{93.83}} & \cellcolor[HTML]{EFEFEF}{\textbf{92.37}}    & {\textbf{96.8}}   & \textbf{94.35}  & \textbf{92.68}  &\textbf{97.25}  &\textbf{95.7}   &\textbf{94.99}   \\

\multirow{-5}{*}{\tabincell{c}{CLIP$-\text{ViT-B/16}$}} & {ETU}   & \cellcolor[HTML]{EFEFEF}{88.47}   & \cellcolor[HTML]{EFEFEF}{78.61}  & \cellcolor[HTML]{EFEFEF}{73.07}  &\cellcolor[HTML]{EFEFEF}{92.69}  &\cellcolor[HTML]{EFEFEF}{87.5} & \cellcolor[HTML]{EFEFEF}{84.89}   & {93.55}   & 89.01  & 85.88   &94.25  &91.3  &89.31   \\\hline

{} & {UniA}    & {32.82}     & 19.34    & {14.42} &   43.77 & {26.3} & 19.31    & {58.85}   & 43.32  & 36.91   & 64.29  & 47.74 & 40.59   \\
{} & {MulA}  & {42.53}  & 25.58  & {20.19}  &   48.37 & {29.74} & 22.63    & {66.73}   & 51.81  & 44.21    & 69.5  & 53.13 & 45.62   \\
{}  & {ETU$_\text{L}$}  & {46.74}  & 28.01   & {21.94} & 50.19 & {31.47} & 23.99   & {71.27}   & 56.16  & 49.6 & 70.89  & 55.94  & 48.9   \\
{}  &{ETU$_\text{S}$}  & {47.25}  & 30.76   &{24.1}  & 55.61 & {39.13} & 32.1    & {72.21}   & 58.58  & 51.71 & 74.28 & 60.89 & 54.49  \\
\multirow{-5}{*}{\tabincell{c}{CLIP$-\text{ResNet50}$}} & {ETU}  & {\textbf{56.83}}    & \textbf{38.9}  & {\textbf{32.54}} & \textbf{61.27} & {\textbf{43.11}} & \textbf{36.59}  & {\textbf{78.55}}   & \textbf{66.85}  & \textbf{61.13}  &\textbf{ 79.37} &  \textbf{67.18} & \textbf{60.72}    \\\hline

{} & {UniA}  & {28.35}     & 13.64    & {9.17}  & 33.93 & {18.17} & 14.14     & {50.27}   & 35.14  & 29.22   & 54.32  & 39.79 & 33.39   \\
{} & {MulA}   & {35.89}     & 18.08    & {13.08}  & 38.46 & {22.42} & 17.03     & {59.13}   & 44.03  & 37.39  & 60.51  & 45.84 & 39.52   \\
{}  & {ETU$_\text{L}$}  & {37.29}  & 20.82   & {14.01} & 40.99 & {24.62} & 17.62    & {61.59}   & 47.43  & 40.72 & 62.03 & 47.91  & 41.33   \\
{}  &{ETU$_\text{S}$}   & {41.51}  &24.0  &{17.82}  & 48.51 & {32.34} &26.25   & {65.47}   & 52.43  & 45.79 & 68.71 & 55.13 & 49.19 \\
\multirow{-5}{*}{\tabincell{c}{CLIP$-\text{ResNet101}$}}   & {ETU} & {\textbf{52.49}}    & \textbf{33.83}  & {\textbf{26.57}} & \textbf{54.27} & {\textbf{37.29}} & \textbf{30.29}   & {\textbf{72.01}}   & \textbf{60.05}  & \textbf{53.8}  & \textbf{73.59} & \textbf{60.79} & \textbf{54.4}   \\\hline

{} & {UniA}   & {19.75}     & 6.85    & {3.35}   & 29.16 & {14.41} & 9.27     & {40.75}   & 23.6  & 17.56  & 46.39  & 28.98 & 23.04 \\
{} & {MulA}   & {21.84}   & 7.13  & 4.47   & 31.12  &14.97 & 10.05     & {41.32}   & 24.18  & 18.67   &47.08  &30.48   &24.47 \\
{}  & {ETU$_\text{L}$} & {22.21}   & 7.68  & 4.07    &31.83  &16.02 & 10.49    & {43.49}   & 26.74  & 21.07  &49.22  &32.41   &25.98   \\
{}  &{ETU$_\text{S}$}   & {20.74}   & 7.54  & 4.78   &32.28  &15.95 & 10.95    & {42.92}   & 25.98  & 20.16   &49.27  &32.04   &25.99 \\
\multirow{-5}{*}{\tabincell{c}{CLIP$-\text{ViT-B/32}$}} & {ETU}   & {\textbf{22.58}}   & \textbf{7.89}  & \textbf{5.39}   &\textbf{33.7}  &\textbf{16.35} & \textbf{11.4}    & {\textbf{45.14}}   & \textbf{27.5}  & \textbf{21.98}  &\textbf{50.51}  &\textbf{33.13}  &\textbf{26.82}  \\\hline

{} & {UniA}   & {14.48}     & 4.57    & {2.54}  & 21.81 & {8.99} & 5.91    & {35.22}   & 20.96  & 16.61  & 39.22  & 24.41 & 19.26   \\
{} & {MulA} & {20.6}   & 8.93  & 5.28   &27.93  &13.9   &10.25    & {43.42}   &29.49  &\textbf{24.16}   &45.93 &31.33  &25.89  \\
{}  & {ETU$_\text{L}$}  & {16.69}   & 5.3  & 2.7    &22.39  &8.81  &6.02   & {38.23}   &23.36  &18.19   &38.72 &23.99  &18.83   \\
{}  &{ETU$_\text{S}$} & {19.26}   & 7.48  &4.78  &\textbf{33.76}  &\textbf{18.48}  &\textbf{13.96}    & {\textbf{45.21}}   &\textbf{29.57}  &{23.19}  &\textbf{53.38 }&\textbf{39.15}  &\textbf{33.54}   \\
\multirow{-5}{*}{\tabincell{c}{CLIP$-\text{ViT-L/14}$}}  & {ETU}  & {\textbf{20.86}}   &\textbf{9.1}  &\textbf{5.49}  &28.54  &14.34  &9.25    & {43.49}   &28.31  &22.77   &47.35 &32.31  &26.56  \\\hline

{} & {UniA}   & {6.36}  & 2.1 & {1.3}  & 11.3 & {3.79} & 2.12  & {20.8}  & 8.7  & 5.41 & 21.86  & 10.37 & 6.95   \\
{} & {MulA}   & {8.86}   &3.11  & 1.9  & 13.52 & 4.8  & 3.09 & {25.93}   &11.75  & 7.64 & 24.92 & 12.36  & 8.22   \\
{}  & {ETU$_\text{L}$}  & {10.11}   &4.01  & 2.8  & 15.32 & 5.7  & 3.36  & {29.21}   &14.32  & 9.57 & 27.79 & 14.35  & 10.28    \\
{}  &{ETU$_\text{S}$}  & {8.76}   &3.21 & 1.9 & 13.26 & 4.12 & 2.71 & {24.65}   & 12.2  & 8.6 & 25.58 & 12.92 & 9.22  \\
\multirow{-5}{*}{\tabincell{c}{ALBEF}}  & {ETU} & {\textbf{13.14}}   &\textbf{4.81} & \textbf{3.3}  & \textbf{17.28} & \textbf{6.54} & \textbf{4.21}     & {\textbf{32.43}}   &\textbf{15.76}  & \textbf{11.28} & \textbf{29.82 }& \textbf{16.25}  & \textbf{11.83}  \\\hline

{} & {UniA}   & {8.54}     & 2.31    & {1.3} & 13.71 & {4.49} & 2.7   & {22.41}   & 9.83  & 6.41  & 23.05  & 11.12 & 7.55   \\
{} & {MulA}   & {13.38}   &4.5 & 2.6  & 17.19 & 6.02 & 3.88   & {27.28}   & 13.02  & 8.42  & {25.68}   & 13.06 & 9.12    \\
{}  & {ETU$_\text{L}$}  & {14.65}   &6.03 & 3.81  & 19.48 & 6.97 & 4.59   & {31.4}   &15.37 & 10.18 & 27.26 & 14.37 & 9.87  \\
{}  &{ETU$_\text{S}$}  & {12.43}  &4.12  &{3.01}  & 17.33 & {6.49} &4.06   & {27.14}   & 13.54  & 9.37 & 26.5 & 13.49 & 9.42 \\
\multirow{-5}{*}{\tabincell{c}{TCL}} & {ETU}  & {\textbf{18.55}} & \textbf{7.94}  & {\textbf{5.71}}   & \textbf{21.57} & {\textbf{8.64}} & \textbf{5.91}  & {\textbf{34.02}}   &\textbf{17.57} & \textbf{11.85}  & \textbf{30.28} & \textbf{15.89} & \textbf{11.23}  \\\toprule 
\end{tabular}
\end{table*}

\noindent\textbf{Downstream tasks and datasets.} 
To comprehensively evaluate the performance of the proposed method, we conduct experiments on three vision-language tasks with three datasets. The first is the image-text retrieval task. Two widely used datasets, i.e., Flickr30K \cite{plummer2015flickr30k} and MSCOCO \cite{lin2014microsoft}, are selected. The Flickr30K data consists of 31,783 images, with each image accompanied by five descriptive captions. The MSCOCO dataset released in 2014 is composed of 164k images, each annotated with approximately five captions. The second is the image captioning tasks, where Flickr30K and MSCOCO are used for the training and testing in the experiments. The third is the visual grounding task. The Flickr30K is used for training. And we choose the RefCOCO+ dataset \cite{yu2016modeling} for the performance evaluation, which contains 141,564 expressions corresponding to 49,856 objects found in 19,992 images. For all experiments, in line with previous research \cite{zhang2022towards,lu2023set}, we use the test set in these datasets for both training and testing purposes. 

\noindent\textbf{Models.} Four widely-used VLP models are utilized to test the proposed method, i.e.,  CLIP \cite{radford2021learning}, ALBEF \cite{li2021align}, TCL \cite{yang2022vision} and BLIP \cite{li2022blip}.  For CLIP, different image encoders are utilized, including vision transformers (i.e., ViT-B/16,  ViT-B/32, and  ViT-L/14\cite{dosovitskiy2020image}) and CNNs (i.e., ResNet50, and ResNet101 \cite{he2016deep}). The text encoder is a 6-layer transformer. For BLIP, we choose the one that consists of a ViT-B/16 and a 6-layer transformer as the image and text encoder to attack. In addition, BLIP is mainly used for experiments on the image captioning task. ALBEF and TCL take ViT-B/16 as the image encoder and adopt a 6-layer transformer for both the text encoder and multimodal encoder. 

\noindent\textbf{Evaluation metric.} In line with prior research \cite{zhang2022towards,lu2023set}, we utilize the Attack Success Rate (ASR) as a metric to quantify the effectiveness of the proposed attack and all compared baselines. ASR is calculated as the percentage of adversarial examples that successfully deceive the model, providing a reliable measure of the attackers' effectiveness. 

\noindent\textbf{Implementation details.} For fundamental experiments, the perturbation magnitude $\epsilon$ is uniformly set as 12/255. Additionally, we evaluate the proposed method under varying perturbation magnitudes. PGD is utilized to solve the optimization problem, with the number of iterations $T$ as 100 and the step size as $\epsilon/T *1.25$. The batch size is set as 16. $\alpha = 4, \beta_1=0.8, \beta_2=0.2$. For $\ell$, we leverage the KL-divergence loss to measure the difference between two samples.

\begin{table*}[!t]
\center
\caption{Attack success rate $(\%)$ regarding the average of R@1 on the image-text retrieval task. The grey background indicates the white-box attack results. \textbf{Bold} indicates the best results.}
\label{tab_cross}
\centering
\resizebox{17cm}{!}{
\begin{tabular}{@{}c|c|cc|cc|cc|cc|cc|cc|cc@{}}
\toprule
\multicolumn{2}{c|}{Test Dataset}  &\multicolumn{14}{c}{Flickr30K}  \\ \hline
\multicolumn{2}{c|}{\multirow{2}{*}{Target Model}} & \multicolumn{10}{c|}{CLIP}  &\multicolumn{2}{c|} {\multirow{2}{*}{ALBEF}} & \multicolumn{2}{c}{\multirow{2}{*}{TCL}} \\ \cline{3-12} 
 \multicolumn{2}{c|}{}& \multicolumn{2}{c|}{\text{ResNet50}}  &\multicolumn{2}{c|}{\text{ResNet101}}  & \multicolumn{2}{c|}{$\text{ViT-B/16}$}   & \multicolumn{2}{c|}{$\text{ViT-B/32}$}   & \multicolumn{2}{c|}{$\text{ViT-L/14}$}  &\multicolumn{2}{c|} {} & \multicolumn{2}{c}{} \\ \hline
{Source Model}   & { Method} & I2T & T2I & I2T & T2I  & I2T & T2I & I2T & T2I & I2T & T2I & I2T & T2I & I2T & T2I \\ \hline
{} & {UniA}   & \cellcolor[HTML]{EFEFEF}{{94.13}}  & \cellcolor[HTML]{EFEFEF}{96.78}    & {16.86} & 21.2      & {8.22}  & 14.5   & \textbf{17.79}  & 25.97    & 7.85    & {16.98}   & 3.86   & {9.33}   & 8.54  & {13.71} \\
{} & {MulA}    &\cellcolor[HTML]{EFEFEF}{96.3}  &\cellcolor[HTML]{EFEFEF}{97.67}   & {18.9}  &24.36 & 8.34   & 13.85   & 16.56    &25.04     & 8.83   & {16.3}   & 5.11    & {8.49}  & 13.38& {17.19} \\
{}  & {ETU$_\text{L}$}   &\cellcolor[HTML]{EFEFEF}{95.79} &\cellcolor[HTML]{EFEFEF}{97.8}    & {27.08}  &34.44 & 7.48  & 16.72   & 14.72  &25.81     & 9.45   & {18.27}   & 4.1     & {9.29} & 14.65& {19.48}  \\
{}  &{ETU$_\text{S}$}  &\cellcolor[HTML]{EFEFEF}{\textbf{96.81}}  &\cellcolor[HTML]{EFEFEF}{\textbf{98.66}}  & {28.22} &34.55   &\textbf{9.69}    & 15.82  & 15.54    &24.58     & \textbf{10.43}   & \textbf{19.46}    & 4.8   & {9.57}  & 12.43& {17.33} \\
\multirow{-5}{*}{\tabincell{c}{CLIP-\text{ResNet50}}}    & {ETU}  &\cellcolor[HTML]{EFEFEF}{95.79} &\cellcolor[HTML]{EFEFEF}{98.63}  & {\textbf{37.04}} &\textbf{44.94}    &9.33     &\textbf{17.49} & {16.58 }   &\textbf{27.74}    & 9.82  & {18.65}    & \textbf{5.42}   & {\textbf{10.59}} & \textbf{18.55}& {\textbf{21.57}}  \\\hline

{} & {UniA} & {18.65}   & 27.65   & {12.77}  & 17.5    & {7.36}   & 13.11   & 17    & 23.36   & 9.08  & {18.4}     & \cellcolor[HTML]{EFEFEF}{79.35}     & \cellcolor[HTML]{EFEFEF}{85.85} & 20.76 & {22.38}  \\
{} & {MulA}  & 19.16  & 28.87  & {11.88}  & 16.5   & 7.12   &12.56 & 15.91   & 25.26     & 9.2   & {17.53}  & \cellcolor[HTML]{EFEFEF}{82.33}    & \cellcolor[HTML]{EFEFEF}{\textbf{87.35}}    & 25.29 & {27.0} \\
{}  & {ETU$_\text{L}$}  & \textbf{31.33}  & 31.05  & {15.07}  & 20.17  & 9.71    &15.14 & 17.06  & 25.29     & 11.78   & {19.68}    & \cellcolor[HTML]{EFEFEF} {79.98}    & \cellcolor[HTML]{EFEFEF}{85.97}  & 21.75& {26.45} \\
{}  &{ETU$_\text{S}$} & 18.14  & 25.28  & {11.37} & 17.26  & 6.99   &12.79 & 16.69  & 23.71  & 8.1   & {15.59}      & \cellcolor[HTML]{EFEFEF}{74.87}   & \cellcolor[HTML]{EFEFEF}{72.99}   & 18.65& {19.62} \\
\multicolumn{1}{c|}{\multirow{-6}{*}{ALBEF}}  & {ETU$_\text{LS}$}  & {26.56}  & \textbf{35.03} & {\textbf{20.31}} & \textbf{25.08} & \textbf{10.06}   &\textbf{17.72}  & \textbf{17.18}  & \textbf{26.22}   & \textbf{13.01}    & {\textbf{21.55}}      & \cellcolor[HTML]{EFEFEF}{\textbf{83.0}}   & \cellcolor[HTML]{EFEFEF}{87.07}   & \textbf{32.98}& {\textbf{31.5}}\\\hline

{} & {UniA}  & {17.58} & 32.08    & {14.18}   & 21.2  & {8.22}   & 14.2  & 17.42   & 26.48    & 10.18    & {18.46}    & 21.58    & {22.57}  & \cellcolor[HTML]{EFEFEF}{90.38} & \cellcolor[HTML]{EFEFEF}{82.26}  \\
{} & {MulA} & 17.31   & 29.16  & {14.3} & 18.94 & 7.26   &13.72    & 14.85   & {25.77}      & 10.55    & {17.56}   & 17.52    & {21.75}   & \cellcolor[HTML]{EFEFEF}{93.68}& \cellcolor[HTML]{EFEFEF}{87.98} \\
{}  & {ETU$_\text{L}$} & 24.27   & 35.68& {17.62} & 22.74  & 8.22   &16.33  & 17.18   & 28.16  & 11.41    & {19.78}     & 23.15   & {24.58}    & \cellcolor[HTML]{EFEFEF}{93.72}& \cellcolor[HTML]{EFEFEF}{\textbf{88.1}} \\
{}  &{ETU}  & {20.56}  & 30.15 & {13.67}  &19.76 &{7.48}  &15.17 & 16.07  & 26.68     & 8.59   & {18.07}   & 18.87   & {20.65}   & \cellcolor[HTML]{EFEFEF}{87.67} & \cellcolor[HTML]{EFEFEF}{83.64} \\
\multirow{-5}{*}{\tabincell{c}{TCL}} & {ETU$_\text{LS}$}  & {\textbf{27.59}} & \textbf{39.69} & {\textbf{20.82}}  & \textbf{26.96}& {\textbf{9.2}}  & \textbf{17.94}   & \textbf{17.44}    & \textbf{28.61}   & \textbf{12.64}   & {\textbf{20.59}}    &\textbf{23.25}   & {\textbf{25.44}}    &\cellcolor[HTML]{EFEFEF}{\textbf{94.1}} & \cellcolor[HTML]{EFEFEF}{87.6} \\\hline\hline

\multicolumn{2}{c|}{Test Dataset}  &\multicolumn{14}{c}{MSCOCO}  \\ \hline
{} & {UniA}  & {93.75}  & 96.2 & {32.28}  & 37.28    & {21.21}     & 26.31    & 31.61  & 38.77   & 24.38    & {28.82}   & 14.43  & {17.37}     & 17.78  & {20.63} \\
{} & {MulA} &93.83    &95.5  & {33.14}  &38.92   & 21.48    & 25.77  & 29.99    &38.13     & 24.72   & {28.84}   & 14.44    & {16.77}    & 16.98& {20.27}\\
{}  & {ETU$_\text{L}$}  &93.87 &95.07   & {46.26}   &53.08  & 21.79   & 28.15  & 30.71        &38.43    & 28.58  & {32.99}   & 16.65    & {18.68}    & 17.46& {20.97} \\
{}  &{ETU$_\text{S}$} &96.61  &96.92    & {51.98}  &55.84  &24.84   &28.39  & 30.74    &39.04    & 29.11    & {32.34}     &17.35   & {19.45}  & 19.89& {21.77} \\
\multirow{-5}{*}{\tabincell{c}{CLIP-\text{ResNet50}}}   & {ETU} &\textbf{96.64} &\textbf{97.3}    & {\textbf{61.26}}  &\textbf{64.95}  &\textbf{25.79}  &\textbf{31.7}  & \textbf{31.63}     &\textbf{40.27}   & \textbf{30.52}    & {\textbf{33.09}}     & {\textbf{17.99}}   & {\textbf{20}}   & \textbf{20.97} & {\textbf{23.1}}\\\hline

{} & {UniA}  & {37.64}  & 43.3  & {24.52}  & 31.68     & {18.81}     & 23.47  & 29.91   & 38.44    & 27.36   & {30.74}    & 79.71   & {83.18}  & 37.22 & {31.12}  \\
{} & {MulA}  & 37.88   & 44.46  & {25.75}  & 32 & 20.18    &24.28  & 31.1  & 38.15     & 26.17    & {30.89}   & 82.57   & {84.88}   & 45 & {37.58} \\
{}  & {ETU$_\text{L}$} & 44.34   & 49.89   & {31.71}  & 38.41  & 22.87   &27.56  & 31.4   & 40.14  & 32.51    & {35.17}     & 79.82   & {83.34}   & 45.56 & {37.24} \\
{}  &{ETU$_\text{S}$}  & 36.9  & 41.82  & {23.54}  & 30.94& 18.58  &24.37   & 30.87  & 38.2  & 22.36     & {28.2}     & 78.68 & {75.79}   & 32.54& {27.62} \\
\multicolumn{1}{c|}{\multirow{-6}{*}{ALBEF}} & {ETU}  & \textbf{49.9}  & \textbf{56.11}  & {\textbf{40.42}} & \textbf{45.16}& \textbf{25.6}   &\textbf{31.95} & \textbf{33.57}  & \textbf{41.47}      & \textbf{34.38}   & {\textbf{37.32}}  &\textbf{82.99}  & {\textbf{85.19}}    & \textbf{50.9}& {\textbf{42.04}} \\\hline

{} & {UniA}  & {41.64} & 49.62  & {31.19}  & 37.21    & {21.75}      & 27.28  & 32.62  & 41.05  & 30.14    & {32.05}      & 43.57    & {36.87}   & 92.01& {86.87} \\
{} & {MulA}  & 38.46  & 47.08  & {28.16} & 33.58  & 21.33   &26.19    & 30.68    & {38.09}   & 26.78    & {31.21}     & 38.7  & {32.01}   & 92.43& {87.36} \\
{}  & {ETU$_\text{L}$}& 47.32  & 54.57  & {36.04}  & 42.29  & 24.8    &29.74 & 33.31  & 41.69   & 29.11    & {32.26}     & 45.13  & {37.25}   & \textbf{93.89}& {89} \\
{}  &{ETU$_\text{S}$} & {39.4}  & 47.89 & {24.38} &34.33   &{20.03}   &21.75 & 30.83  & 41.35   & 27.05    & {30.69}   & 42.92  & {36.91}     & 92.28& {88.57}\\
\multirow{-5}{*}{\tabincell{c}{TCL}}   & {ETU}  & {\textbf{53.25}} & \textbf{57.99} & {\textbf{42.3}}  & \textbf{46.58}  & {\textbf{26.4}} & \textbf{31.9}   & \textbf{33.88}     & \textbf{41.78}    & \textbf{31.48}   & {\textbf{33.83}}   & \textbf{45.42}   & {\textbf{37.93}}  & {93.28}& {\textbf{89.45}}  \\\toprule 
\end{tabular}}
\end{table*}

\noindent\textbf{Baselines.} As the proposed method is the first endeavor to learn UAPs against VLP models, there are no existing benchmarks for comparison. To address this, we construct baselines based on prior works that focus on sample-specific attacks against VLP models \cite{zhang2022towards}, along with variants of our proposed method. These can help verify the efficacy of each component and demonstrate the superiority of our overall algorithm in learning UAPs.

\noindent(1) Unimodal attack (UniA), which learns UAPs by directly enlarging the embedding distance between the adversarial images and their original counterparts; 

\noindent(2) Multi-modal attack (MulA), which pulls adversarial images away from the original images and paired texts in the embedding space. The objective is $\max _{\delta} \mathcal{L}_{1}$;

\noindent(3) ETU$_\text{L}$, which is a variant of the ETU that only considers local utility reinforcement. The objective is $\max _{\delta} (\mathcal{L}_{1} + \mathcal{L}_{2})$;

\noindent(4) ETU$_\text{S}$, which is a variant of the proposed method that utilizes the ScMix augmentation to increase the input diversity. The augmented data are utilized to optimize the entire area of UAPs without considering the local utility.

\subsection{Results on the Image-Text Retrieval}
We assess the transferability of the proposed method by launching attacks against a range of VLP models, i.e., CLIP models with different backbones, ALBEF and TCL, in black-box settings. Given that different VLP models may accept inputs of varying sizes, the learned UAPs are resized accordingly before initiating attacks. For instance, the UAPs learned based on CLIP would be resized to $384 \times 384$ when attacking ALBEF or TCL. Conversely, when conducting transfer attacks from ALBEF or TCL to CLIP, the UAPs are resized from $384 \times 384$ to $224 \times 224$.

Table \ref{tab_trans_1} presents the attack results by taking $\text{ViT-B/16}$-based CLIP and the Flickr30K dataset as the source model and training set, respectively. The averages of R@1, R@5, and R@10 for both image-to-text retrieval and text-to-image retrieval are recorded. From these results, we can draw the following observations.

\begin{table}[!t]
\caption{Performance under different attacks on the visual grounding task. The training dataset and test dataset are Flickr30K and RefCOCO+, respectively. The CLIP with the ViT-B/16 is set as the source model and ALBEF is taken as the target model. The ``Baseline'' denotes the performance of the target model on the original data. Lower values represent better adversarial transferability.  \textbf{Bold} indicates the best results.} 
\label{tab_trans_2}
\centering
\begin{tabular}{@{}cc|cccccc@{}}
\toprule
& \multicolumn{1}{c|}{Test Dataset}   & \multicolumn{3}{c}{RefCOCO+}  \\ \hline
 & { Method} & {Val  \ \ \ } & {TestA \ \ \ }   & {TestB \ \ \ }   \\ \hline
& {Baseline}   & {51.2 \ \ \ }  &{56.7  \ \ \ } &{44.8 \ \ \ }    \\
& {UniA}   &{49.6 \ \ \ }  &{53.4 \ \ \ }  & {42.9 \ \ \ }    \\
& {MulA}   &{49.6 \ \ \ }  &{53.1 \ \ \ }  & {42.6 \ \ \ }    \\
& {ETU$_\text{L}$}   &{48.7 \ \ \ }  &{52.4  \ \ \ } & {42.1  \ \ \ }  \\
& {ETU$_\text{S}$}   &{49.2 \ \ \ }  &{53.1 \ \ \ }  & {42.5  \ \ \ }  \\
& {ETU}  &{\textbf{48.5} \ \ \ }  &{\textbf{51.5} \ \ \ }  & {\textbf{41.7} \ \ \ } \\\toprule
\end{tabular}
\end{table}

First, all methods achieve promising white-box attack performance. However, when transferring to unseen models, methods that do not consider transferability, i.e., UniA and MulA, lose their efficacy. This means the two methods overfit the source model during training. In contrast, by considering the local utility and multi-modal data diversity, the proposed method can significantly improve black-box attack performance. Interestingly, increasing the diversity of the input to optimize the entire area of UAPs, i.e., ETU$_\text{S}$, would improve attack performance on CLIP while harming the transferability on other models, i.e., ALBEF, and TCL. The reason may be that UAPs are learned based on the whole dataset. Increasing the number of inputs would make UAPs overfit the source model. This might not influence the transferability across models with the same learning objective but influences performance on models with different learning objectives. Second, compared to UniA, MulA achieves better performance, which further demonstrates the necessity of considering cross-model interactions when implementing multi-modal attacks. Third, different architectures show different levels of robustness to UAPs. The models with ViT backbones show stronger robustness to the attack. ALEBF and TCL are less sensitive to the universal adversarial attack that targets CLIP. The reason may be that they adopt different learning objectives and architectures, thus introducing different representation spaces and multi-modal interactions. 

We further test the effectiveness of the proposed method by adopting different source models, where the results are summarized in Table \ref{tab_cross}. The promising performance achieved consistently demonstrates the superior performance of our approach in generating effective and transferable universal adversarial perturbations.

\begin{table}[!t]
\caption{Performance under different attacks on image captioning. The training dataset and test dataset are Flickr30K and MSCOCO, respectively. The CLIP with the ViT-B/16 is adopted as the source model and BLIP is taken as the target model. ``Baseline'' denotes the performance of the target model on original data. Lower values represent better adversarial transferability. \textbf{Bold} indicates the best results.}
\label{tab_trans_3}
\centering
\resizebox{8.5cm}{!}{
\begin{tabular}{@{}cc|cccccccc@{}}
\toprule  
& \multicolumn{1}{c|}{Test Dataset}   & \multicolumn{5}{c}{MSCOCO}  \\ \hline
& { Method} & B@4  & {METEOR}   & {ROUGE$\_\text{L}$}  & {CIDEr} & {SPICE}  \\ \hline
& {Baseline}   & 39.3  &30.7 &59.6   &131.4 &23.5   \\
& {UniA}   & 36.8  &29.3 &57.6   &122.3  &22.0   \\
& {MulA}   & 35.6  &28.7 &56.9   &118.1 &21.6   \\
& {ETU$_\text{L}$} & 35.5  &28.4 &56.6   &117.1 &21.4   \\
& {ETU$_\text{S}$} & 36.0  &28.8 &57.1   &119.4 &21.7   \\
& {ETU}  & \textbf{34.8} &\textbf{28.1} &\textbf{56.3}   &\textbf{114.8} &\textbf{21.0} \\\toprule 
\end{tabular}}
\end{table}

\begin{table}[!t]
\center
\caption{Comparison with different augmentation methods. Attack success rate $(\%)$ on image-text retrieval task are reported. The CLIP with the ViT-B/16 and Flickr30K are adopted as the source model and dataset for training.}
\label{tab_aug_0}
\centering
\resizebox{8.5cm}{!}{
\begin{tabular}{@{}cc|ccc|ccc@{}}
\toprule
\multicolumn{2}{c|}{Test Dataset}   & \multicolumn{6}{c}{MSCOCO}  \\ \cline{0-7}
\multicolumn{2}{c|}{Task} & \multicolumn{3}{c|}{Image-to-Text}     & \multicolumn{3}{c}{Text-to-Image} \\ \hline
{Target Model} & {Method} & R@1  & R@5 & R@10  & R@1  & R@5  & R@10  \\ \hline
{}  & {ETU$_\text{SI}$} & {68.38}   &54.72  & 45.74  & 71.24 & 53.74  & 45.08   \\
{}  & {ETU$_\text{SM}$}  & {73.85}   &62.14  & 55.37  & 76.33 & 63.85  & 51.17  \\
{} &{ETU$_\text{Admix}$}  & {75.56}   & 64.21  & 56.96  & 77.24 & 66.18 & 56.1   \\
\multirow{-4}{*}{\tabincell{c}{CLIP$-\text{ResNet50}$}} &{ETU}  & {\textbf{78.55}}   & \textbf{66.85}  & \textbf{61.13} &\textbf{79.37} &  \textbf{67.18} & \textbf{60.72}    \\\hline
{}  & {ETU$_\text{SI}$}  & {39.11}   &22.79  & 17.39  & 46.18 & 28.88  & 22.47  \\
{}  & {ETU$_\text{SM}$}   & {42.2}   &27.32  & 20.71  & 48.4 & 31.61  & 25.56 \\
{} &{ETU$_\text{Admix}$} & {44.14}   & 27.29  & 21.27   & 50.03 & 32.66 & 26.34   \\
\multirow{-4}{*}{\tabincell{c}{CLIP$-\text{ViT-B/32}$}} &{ETU}  & {\textbf{45.14}}   & \textbf{27.5}  & \textbf{21.98}  &\textbf{50.51}  &\textbf{33.13}  &\textbf{26.82}  \\\toprule 
\end{tabular}}
\end{table}

\begin{figure}[tbh]
\begin{subfigure}[Image-to-Text]
{\includegraphics[angle=0, width=0.23\textwidth]{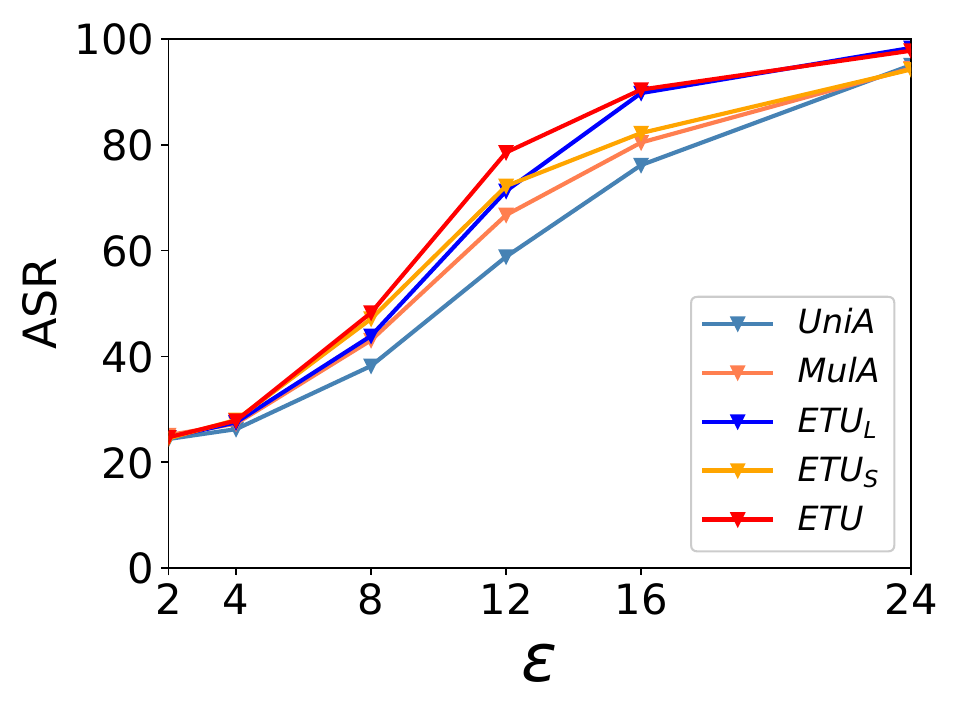}}
\end{subfigure}
\begin{subfigure}[Text-to-Image]
{\includegraphics[angle=0, width=0.23\textwidth]{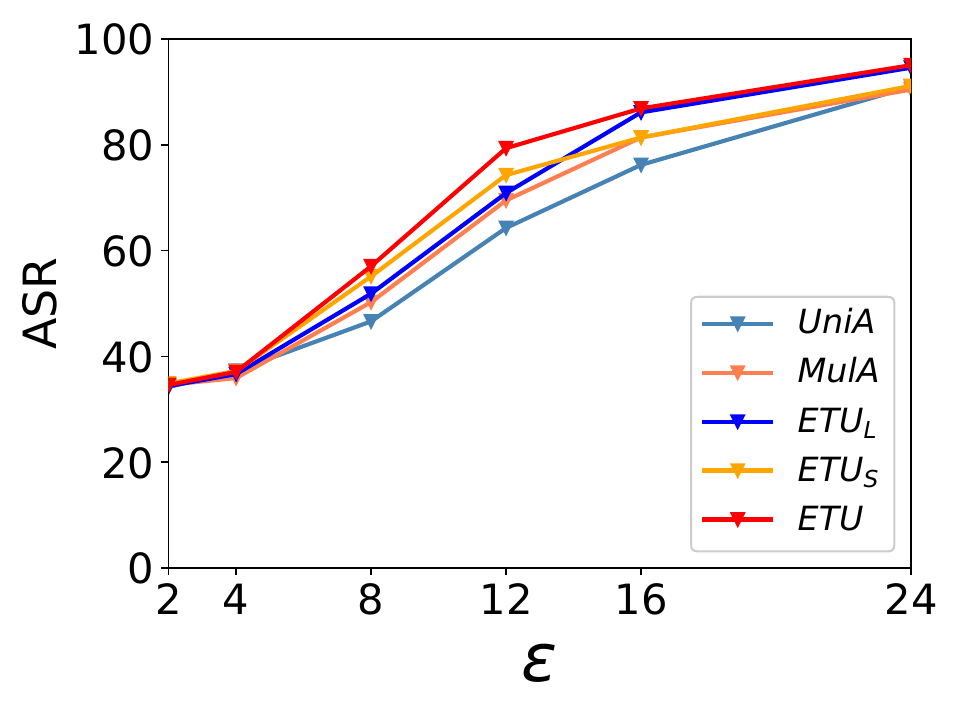}}
\end{subfigure}
\caption{Test accuracy on MSCOCO under different magnitudes of the UAP. The source model is ViT-B/16-based CLIP and the target model is ResNet50-based CLIP. The attack success rate in terms of the average of R@1 is reported.}
\label{datamag_acc}
\end{figure}

\subsection{Results on the Visual Grounding and Image Captioning}

To test the transferability of the proposed method across different tasks, we propose to conduct experiments on the visual grounding task and the image captioning task. For two tasks, we use the UAP generated based on Flickr30K and ViT-B/16-based CLIP in the image-text retrieval to attack target models. For the visual grounding task, we take ALBEF and RefCOCO+ as the target model and dataset. While on the image captioning, we use the learned UAP to attack BLIP on MSCOCO. The performance of target models under different attacks is reported in Table \ref{tab_trans_2} and \ref{tab_trans_3}. From the results, we can have the consistent observation that the proposed method achieves the best universal adversarial attack. In addition, each component contributes to the effectiveness and transferability of the proposed method.

\begin{figure*}[tbh]
\begin{subfigure}
{\includegraphics[angle=0, width=0.97\textwidth]{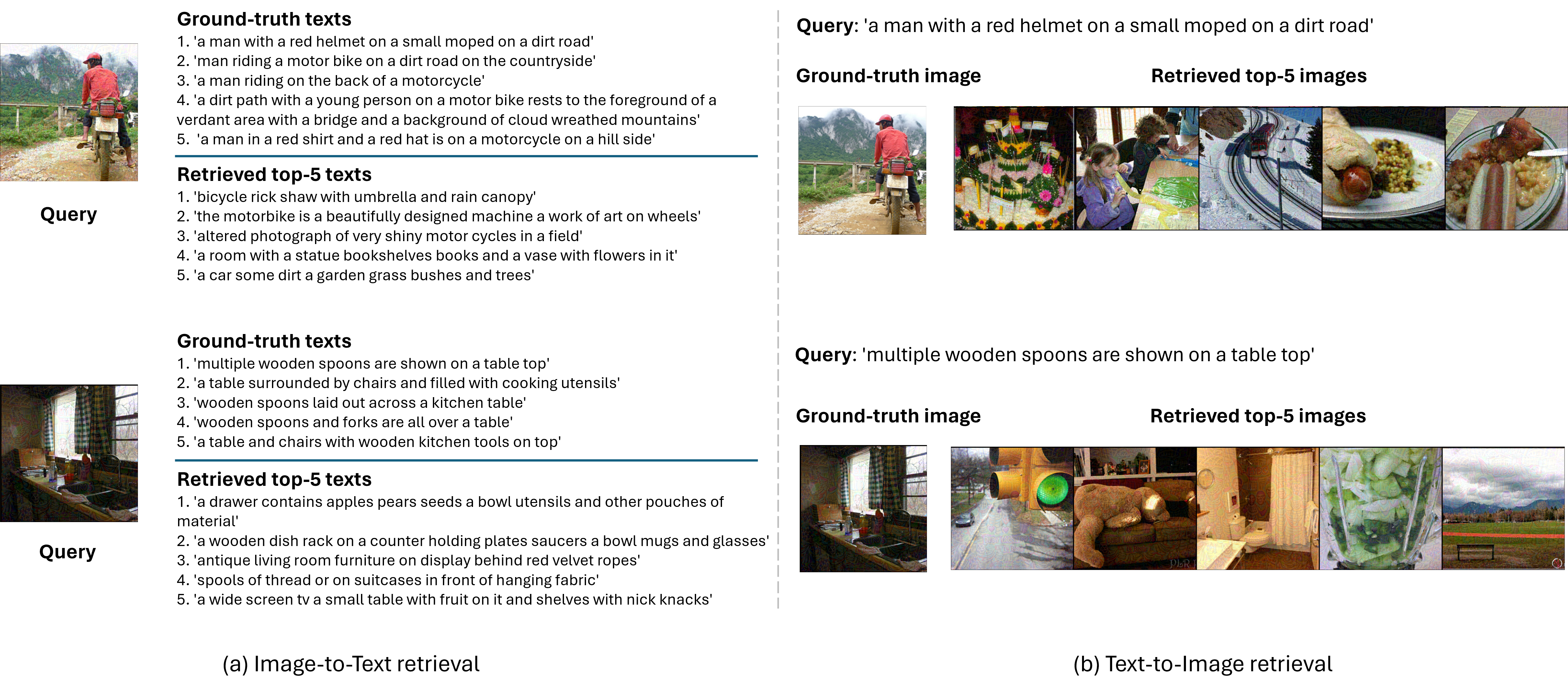}}
\end{subfigure}
\caption{Examples of top-5 image-text retrieval results.}
\label{ITR}
\end{figure*}

\begin{figure}[tbh]
{\includegraphics[angle=0, width=0.42\textwidth]{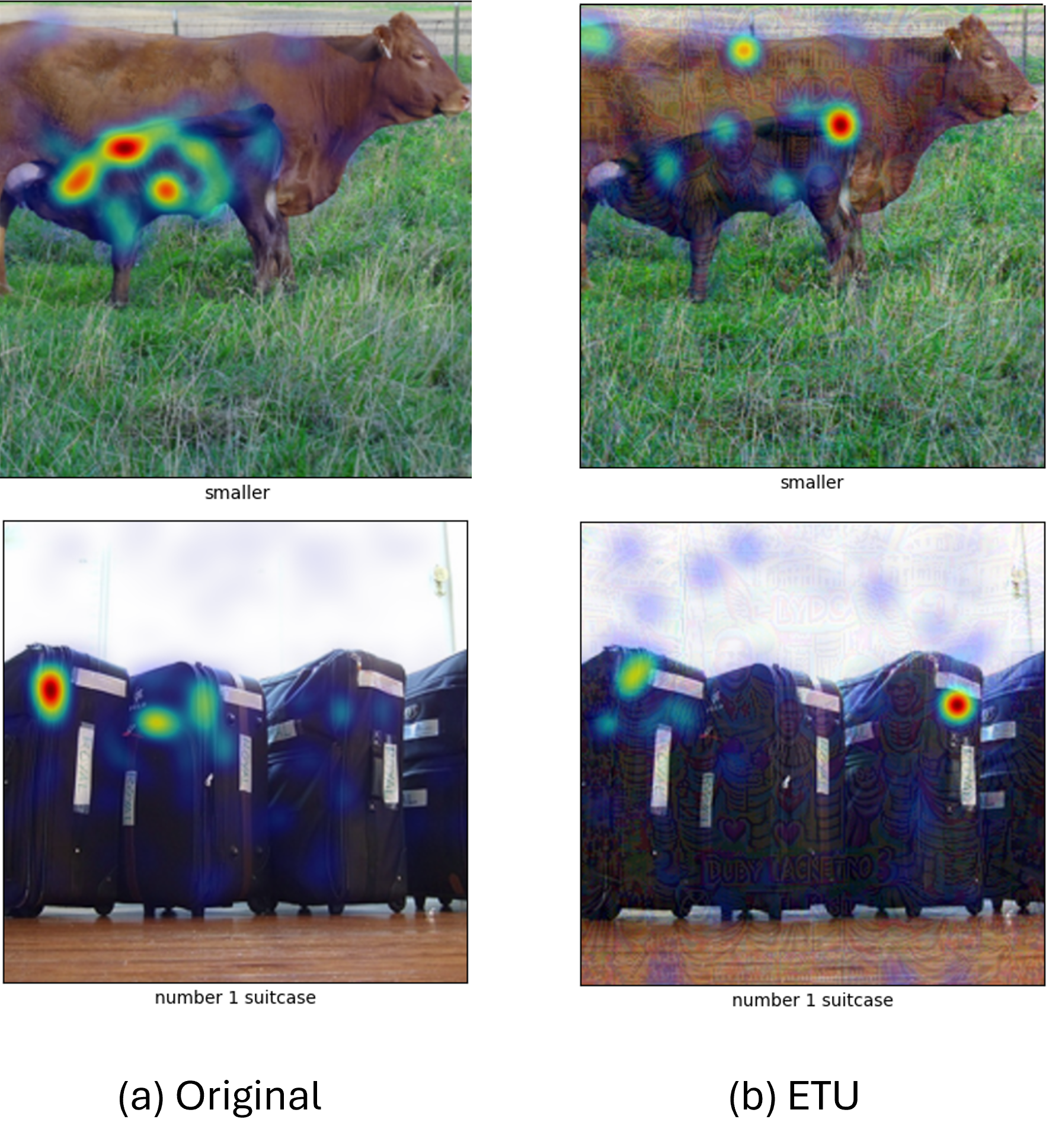}}
\caption{The Grad-CAM visualizations of the original images and the perturbed images by ETU.}
\label{Grad}
\end{figure}

\subsection{Comparison with Different Augmentations}
We compare the proposed ScMix with various data augmentation methods, including scale-invariant augmentation  ($\text{ETU}\_${SI}) \cite{lu2023set}, self-mix ($\text{ETU}\_${SM}), and Admix ($\text{ETU}\_${Admix})  \cite{wang2021admix},  where results on image-text retrieval are reported in Table \ref{tab_aug_0}. From the table, it can be observed that the proposed augmentation method is the most superior.

\balance

\subsection{Results on Varying Perturbation Budgets}
We test all methods under various perturbation magnitudes, with results on image-text retrieval presented in Figure \ref{datamag_acc}. From the figure, we can observe that the proposed method achieves the overall best performance, further confirming its superiority. It is worth noting that as the magnitude increases, the performance of all methods improves. It is quite normal as larger perturbations typically lead to better attack performance. However, large perturbations would significantly influence data quality, making them easier to identify.

\subsection{Visualization}
In Figure \ref{ITR}, we showcase some examples of image-text retrieval under the proposed attack. From the figure, it is evident that the proposed method can effectively mislead the target model to return incorrect retrieved data. Additionally, we present some Grad-CAM \cite{selvaraju2017grad} visualization examples in Figure \ref{Grad}, where it can be seen that the UAP can significantly change the attention of the target model. These further confirm the effectiveness of the proposed method. 

\section{Conclusions}
In this paper, we investigate to learn universal adversarial perturbations that are capable of transferring across different VLP models, datasets and downstream tasks. To this end, we thoroughly study the factors that might influence the utility and transferability of the UAPs. Based on the findings, we propose a novel \underline{E}ffective and \underline{T}ransferable \underline{U}niversal Adversarial Attack (ETU) method. The proposed method achieves effective attacks by comprehensively considering the characteristics of UAPs and complex multi-modal interactions. Local utility enhancement of UAPs and a novel ScMix data augmentation are designed to ensure the performance of the proposed method. We test the proposed methods across different VLP models, downstream tasks and datasets, where promising results demonstrate the superiority of the proposed method. 

\section{Acknowledgments}
This work was partially supported by Australian Research Council Discovery Project (DP230101196, CE200100025).

\clearpage
\bibliographystyle{ACM-Reference-Format}
\balance
\bibliography{acmart}

\end{document}